# Individual Fairness Through Reweighting and Tuning

Abdoul Jalil Djiberou Mahamadou[a*], Lea Goetz[b], Russ Altman[c, d, e, f]

[a]Stanford Center for Biomedical Ethics, Stanford University, Stanford, CA 94305, USA,
[b]Artificial Intelligence and Machine Learning, GSK, London, N1C 4AG, UK
[c]Department of Biomedical Data Science, Stanford University, Stanford, CA 94305, USA
[d]Department of Bioengineering, Stanford University, Stanford, CA 94305, USA
[e]Department of Genetics, Stanford University, Stanford, CA 94305, USA
[f]Department of Medicine, Stanford University, Stanford, CA 94305, USA
*Corresponding Author: abdjiber@stanford.edu

**ABSTRACT**

Inherent bias within society can be amplified and perpetuated by artificial intelligence (AI) systems. To address this issue, a wide range of solutions have been proposed to identify and mitigate bias and enforce fairness for individuals and groups. Recently, Graph Laplacian Regularizer (GLR), a regularization technique from the semi-supervised learning literature has been used as a substitute for the common Lipschitz condition to enhance individual fairness. Notable prior work has shown that enforcing individual fairness through a GLR can improve the transfer learning accuracy of AI models under covariate shifts. However, the prior work defines a GLR on the source and target data combined, implicitly assuming that the target data are available at train time, which might not hold in practice. In this work, we investigated whether defining a GLR independently on the train and target data could maintain similar accuracy. Furthermore, we introduced the *Normalized Fairness Gain* score (NFG) to measure individual fairness by measuring the amount of gained fairness when a GLR is used versus not. We evaluated the new and original methods under NFG, the Prediction Consistency (PC), and traditional classification metrics on the German Credit Approval dataset. The results showed that the two models achieved similar statistical mean performances over five-fold cross-validation. Furthermore, the proposed metric showed that PC scores can be misleading as the scores can be high and statistically similar to fairness-enhanced models while NFG scores are small. This work therefore provides new insights into when a GLR effectively enhances individual fairness and the pitfalls of PC.

## 1 INTRODUCTION

When not carefully monitored, biases within our society can be integrated and amplified by AI, especially in critical systems such as healthcare [6]. Identification and mitigation of such bias have become crucial to the wide adoption of these AI technologies. While the sources of AI bias are well documented [11,24] how to mitigate the bias remains an open question as the is no one-size-fits-all approach. Practical solutions, on the one hand, attempt to identify and mitigate bias by the establishment of norms, regulations, ethical frameworks, and checklists [24]. Technical solutions, on the other hand, have been introduced to address AI bias at different model development stages which comprise three main categories [33]: pre-processing, in-processing, and post-processing techniques. The first category focuses on input data modification such as sensitive attribute suppression, data sampling, and relabeling to attain fair outcomes [28]. The second and third categories, respectively operate through the integration of fairness constraints during the training phase of the models and adjusting the model outcomes. These techniques can further be categorized into two main scopes: group and individual fairness. Group fairness techniques aim to promote similar treatment for distinct groups, e.g., ethnic groups, and individual fairness similar treatment for similar individuals.

A main drawback of group fairness techniques is the assumption of access to sensitive attributes often used to define groups. However, in practice, such attributes may not be available or even prohibited from collection and use in decision-making processes by regulatory restrictions [20]. In addition, an AI model can satisfy group fairness requirements while remaining unfair to individuals [12]. In contrast to group fairness, individual fairness does not explicitly rely on sensitive attributes and can enhance group fairness for certain population subgroups [12]. Recently, individual fairness has been applied in diverse fields including recommender systems [34], online learning [16], graph mining [18], and text mining [2] (see Section 2.3 for a detailed discussion).

To attain individual fairness, a popular criterion is Lipschitz's condition, introduced by [12]. Under this framework, a model is fair if, for any pairs of individuals, the ratio of their pairwise distance and the distance of their outcomes is less than a user-defined threshold. When applied to entire datasets (global condition), Lipschitz's condition often comes with a cost to the accuracy of the model. As a substitute, graph Laplacian regularization (GLR) techniques [7] from the semi-supervised learning literature have been introduced to enforce individual fairness. Specifically, [26] showed that GLR not only satisfies the global and local (relaxed versions) Lipschitz condition but can also improve the accuracy of transfer learning techniques under covariate shifts [22]. The proposed technique, which here we refer to as Individual Fairness through Domain Adaptation (IFDA), defines a GLR on the source and target data combined and implicitly assumes access to the target data is available at training time. This may not hold in practice.

Another line of research in algorithmic fairness is fairness evaluation metrics. While numerous group fairness metrics have been proposed [5], metrics for individual fairness are under-researched. The proposed group fairness metrics have several limitations including access to sensitive variables, contradictions between metrics [32], and the impossibility of optimizing some metrics at the same time [8,19]. For individual fairness, a popular metric is the Prediction Consistency (PC) introduced in [36]. Under this metric, a model is individually fair if its outcomes remain the same when sensitive attribute values are perturbed while leaving other variables unchanged. For instance, changing the ethnic group of an individual from White to Black should yield the same outcomes. Like group fairness metrics, this metric assumes access to sensitive variables that might be unavailable in practice.

The aim of our work is two-fold: first, we extend the IFDA technique by defining a GLR independently for the source and target data, i.e., enforcing individual fairness independently. We compare the performance of the new logistic regression model to IFDA. Second, we investigate whether in-processing fairness techniques, especially under the GLR framework, can reliably enforce individual fairness.

To attain the first aim, we extend the Target-Fair Covariate Shift model (TFCS) [9]. TFCS is a group fairness technique that operates when sensitive attributes are available only in the target domain. The model is trained on the source data with any fairness constraints. At inference time, the model is retrained to enforce group fairness. To minimize distribution shifts between the source and target data, TFCS uses instance reweighting [38], a popular transfer learning technique. While solving a practical issue, the retraining phase of TFCS is quadratic in the size of the training data. As a result, we define a procedure that requires only fine-tuning a model. Moreover, we investigate whether there is any utility to using a regularized fairness technique over a non-regularized one when fine-tuning a model by defining the *Normalized Fairness Gain* score. This metric quantifies the fairness gap when a regularization is used versus not.

We used the German Credit Approval dataset to evaluate the new model referred to as Individual Fairness through Reweighting and Tuning (IFRT) and IFDA under the Normalized Fairness Gain, the Prediction Consistency, traditional classification and fairness metrics: AUC ROC, False Negative Rate, and False Positive Rate. As IFDA and IFRT are extensions of logistic regression models, we used a logistic regression model as a baseline model and compared the performances of the models over a five-fold cross-validation procedure. The results showed that IFDA and IFRT achieved similar performances on all evaluation metrics. For Prediction Consistency, statistical analyses under different sensitive variables showed that the logistic regression model achieved the same performance compared to IFDA and IFRT on some sensitive variables. In addition, the NFG scores showed that IFDA and IFRT can attain high Prediction Consistency performances while the fairness gains are small. This has two implications; the Prediction Consistency metric can be misleading as a model with any fairness constraint such as the logistic regression can achieve similar fairness performances as a fairness-enhanced model such as IFDA and IFRT, and a model might look to enhance individual fairness while in reality, it doesn't.

The remaining sections are organized as follows: in Section 2, we define Lipschitz's condition, GLR, and describe related work on individual fairness; we introduce the proposed model and fairness metric respectively in Sections 3 and 4; Section 5 describes the experiments we carried out and obtained results. We provided all experimental codes carried out in this work and additional results in a companion [repository](#).

## 2 BACKGROUND AND RELATED WORK

This section introduces Lipschitz's condition, graph Laplacian regularization, and recent works on individual fairness.

### 2.1 Notations

We followed similar notations in [9] and defined $X_s = (x_i, y_i) \ \forall \ i \in \{1, \dots, n\}$ with $n$ the number of samples and $x_i \in \mathbb{R}^m$ as the source data and similarly, $X_t \in \mathbb{R}^{m'n'}$ as the target data. Let $Q_{X_s}$ and $P_{X_t}$ be the respective distribution of $X_s$ and $X_t$ and $W = (w_1, w_2, \dots, w_n)$ with $w_i \geq 0 \ \forall i$ be a reweighting vector of the source data to minimize distribution shifts between the source and target data. We define a parametric model $f: X_s \to \Upsilon$ where $\Upsilon$ is the space of outcomes (e.g., probabilities, logits, …) of $f$ and $\hat{y} = f(X; \theta)$ where $\theta$ are the parameters of the model. Finally, we define a similarity metrics on $X_s$ and $\Upsilon$ respectively as $d$ and $D$.

### 2.2 Lipschitz's Condition

Individual fairness relies on the principle that similar individuals should have similar outcomes. A popular approach to enforcing individual fairness is Lipschitz's condition, introduced in [12] and defined as follows:

$f$ is $\tau -$Lipschitz if $\forall \ (x_i, x_j) \in X$:

$$D\left(f(x_i), f(x_j)\right) \leq \tau d(x_i, x_j). \tag{1}$$

Where $\tau$ is a user-defined positive threshold.

Satisfying (1) for all pairs of samples (global Lipschitz condition) can come at the cost of model performance. One can relax this condition by requiring (1) to be satisfied only for a subset of data (local condition), for instance by considering pairs of individuals with a distance less than a certain threshold. As a substitute to (1), [26], introduced GLRs for individual fairness enforcement.

### 2.3 Graph Laplacian

A graph Laplacian is a matrix representation of a graph; each node corresponds to a data instance, and the edges correspond to the relationship (e.g., similarity) between the nodes. This representation has been extensively studied and used in semi-supervised learning [7].

Let $K$ be the pairwise similarity matrix on a dataset $X$ where the distance function is the same as in (1). The (unnormalized) graph Laplacian $L$ is the symmetric matrix defined as

$$L = M - K \qquad (2)$$

where $M$ is the degree matrix given by $M_{ii} = \sum_j K(x_i, x_j)$ if $i = j$ and $M_{ij} = 0$ if $i \neq j$. Other versions of $L$ such as the normalized ($L = M^{\frac{1}{2}} K M^{\frac{1}{2}}$) have been studied in the literature [7]. For simplicity and comparison purposes with IFDA, we used the unnormalized graph Laplacian in the remaining manuscript.

## 2.4 Graph Laplacian Regularization

Taking $L$ in a quadratic form, the graph Laplacian can be used as a regularization constraint. Let $R(f(X)) = \frac{1}{2} f(X)^T L f(X)$ be such a quadratic form where $f(X)$ is a vector of outcomes of a model $f$. It can be shown that [7]:

$$R(f(x)) = \frac{1}{2} \sum_{ij} K(x_i, x_j) \big(f(x_i) - f(x_j)\big)^2. \qquad (3)$$

If $K$ is a decreasing function on $X$ (e.g. $K_{ij} = e^{-\delta d(x_i, x_j)}$ with $\delta$ a temperature parameter [26]), minimizing $R$ is equivalent to satisfying Lipschitz's condition (1) [22,26]. It then follows that equation (3) can be used to enforce individual fairness. [26] used such a regularization to post-process unfair AI outcomes to meet individual fairness. In follow-up work, the authors introduced an in-processing fairness technique [22] where $K$ is defined by the source and target combined. Interestingly, the authors showed that such a strategy can help improve the accuracy of transfer learning techniques. For more details on fairness enforcement with GLRs and additional work, we refer readers to [22,26].

## 2.5 Other Work

There are emerging applications and adaptations of individual fairness in recent years. In [36], individual fairness is defined as how robust a model is under some variable modification. That is, a model is individually fair if its performance is invariant under certain perturbations of the sensitive attributes. This definition led to the introduction of Prediction Consistency as an individual fairness metric. [34] defines individual fairness for recommender systems when similar items receive similar coverage in the recommendations. In the same line of work, [4,10,15,30,31,35] defined other individual fairness definitions for recommender systems, and [27] provided guidelines for their applications. In [16], the authors extended the traditional individual fairness definition [12] to online contextual decision-making and introduced the concepts of fairness across time which requires similar treatment of individuals relative to the past and the future and fairness in hindsight for which fairness is required only at the time a decision is made. [1] introduced identity-independent fairness and time-independent fairness for online systems and [18] used a revised version of GLR to define individual fairness for graph mining. Under this application, two similar nodes in a graph should have similar mining results. [37] introduced an individual fairness pre-processing technique through label flipping while [29] defined individual fairness for representation learning. In [21], confidence intervals and test hypotheses for model fairness for statistical inference were proposed; [2] evaluated individual fairness metrics for text classification and provided fair metric selection guidelines. We refer the readers to [25] for a detailed review of individual fairness (and group fairness) in AI.

# 3 PROPOSED MODEL

We introduce an extension of IDFA [22] and TFCS [9] for individual fairness enforcement for applications where sensitive attributes are not required by the model to enhance fairness. To extend IFDA, we define a GLR for the source and target datasets independently. This allows for training the model without access to the target data as in IFDA. At inference, the GLR on the target set is used to fine-tune the model to enhance individual fairness. Note that this is different from the retraining in TFCS.

Overall, the proposed model, referred to as Individual Fairness through Reweighting and Tuning (IFRT) here can be applied to applications where fairness is desired only on the source data, only on the target data, or both. Furthermore, it can be applied under a covariate shift setting with the instance reweighting mechanism of TFCS. In the following, we describe the theoretical working of IFRT in which we consider a binary classification with SoftMax outcomes for simplicity.

Let $\theta = (\theta_1, \theta_2, \ldots, \theta_m)^T$ be the learned parameters of a model $f$, and $f(X; \theta)$ be the SoftMax outcomes (i.e., $f(X; \theta) = 1/(1 + \exp(-X\theta))$). We define the loss function on the source data by (4) and target data by (5):

$$\mathcal{L}_s(\theta) = \frac{1}{n}\sum_{i=1}^{n} w_i l\big(f(x_i; \theta)\big) + \alpha R_s(f(X_s; \theta))). \tag{4}$$

$$\mathcal{L}_t(\hat{\theta}) = R_t\big(f(X_t; \hat{\theta})\big). \tag{5}$$

Where $l\big(S(X; \theta)\big)$ is a traditional classification loss function such as the cross-entropy, $R_s\big(f(X_s; \theta)\big) = \frac{1}{2} f(X_s; \theta)^T L_s f(X_s; \theta)$ is the GLR on $X_s$, and $R_t$ the GLR on $X_s$; $\alpha \geq 0$ is a user-defined threshold corresponding to the regularization strength, $\hat{\theta} = argmin_\theta \mathcal{L}_s(\theta)$ are the optimal parameters of $f$.

In [22], the authors defined a train loss function similar to (4) with a GLR set to $R_{s+t}$ while in [9], the authors assumed that at train time any regularization is absorbed into $l$. Here, as $R_s$ is a function of $X_s$ and $\theta$, theoretical results presented in [9] partially do not hold. Furthermore, [9] defined an inference loss similar to (4) where the regularizer is the average disparity score obtained from sensitive attributes. A notable difference between IFRT and TFCS is that to adapt to the target data, TFCS requires fine-tuning $W$ which leads to retraining $f$ at inference. We designed (5) to avoid a retraining cost in IFRT by requiring only a fine-tuning of $\hat{\theta}$.

Similar to [9], we used gradient descent to solve the optimization problem. As (4) and (5) are convex, the optimums are global. To solve the optimization problem, we defined $l$ as a binary cross-entropy loss (i.e., $l\big(s_i(x_i; \theta)\big) = -y_i \log\big(f_i(x_i; \theta)\big) - (1 - y_i)\log(1 - f(x_i; \theta))$ similar to [9]. Under this setting and the chain rule, the gradient of (4) and (5) are respectively given by (6) and (7).

$$\nabla_\theta \mathcal{L}_s(\theta) = \nabla_\theta f \nabla_f \mathcal{L}_s(\theta) + \alpha \nabla_\theta f \nabla_f R_s\big(f(X_s; \theta)\big) \tag{6}$$

$$= \frac{1}{n}\sum_{i=1}^{n} w_i(f_i(X_s; \theta) - y_i)x_i + \alpha G_s(\theta),$$

Where $G_s(\theta) = X_s J_s L_s f(X_s; \theta)$, with $J_s = diag[f(X_s; \theta)(1 - f(X_s; \theta))]$ and

$$\nabla_{\hat{\theta}} \mathcal{L}_t(\hat{\theta}) = G_t(\hat{\theta}). \tag{7}$$

Given (6) and (7), the IFRT train and inference algorithms are provided by ALGORITHM 1 and ALGORITHM 2.

---

**ALGORITHM 1: Train**

**Inputs**: $X_s$ the train data (to account for intercepts in $\theta$, a one-vector column can be added to $X_s$), $y_s$ the source labels, $\alpha \geq 0$ the trade-off threshold, $\eta_\theta > 0$ the learning rate of $\theta$,
$W$ the instance reweighting vector which can be uniform under IID and covariate shift weights [9] under changing environments.
$\theta \leftarrow$ random or zero initialization;
$k \leftarrow 1$;
**Repeat**
  Get $\nabla_\theta \mathcal{L}_s(\theta_k)$ from (6);
  Set $\theta_{k+1} \leftarrow \theta_k - \eta_\theta \nabla_\theta \mathcal{L}_s(\theta_k)$;
  $k \leftarrow k + 1$;
**Until** Convergence
**Outputs**: $\hat{\theta}$

---

**ALGORITHM 2: Inference**

**Inputs**: $X_t$ the target data, $\hat{\theta}$ the optimal parameters from ALGORITHM 1, $\eta_{\hat{\theta}} > 0$ the fine-tuning learning rate.
$k' \leftarrow 1$;
**Repeat**
  Get $\nabla_{\hat{\theta}} \mathcal{L}_t(\widehat{\theta_{k'}})$ from (7);
  Set $\widehat{\theta}_{k'+1} \leftarrow \widehat{\theta}_{k'} - \eta_{\hat{\theta}} \nabla_{\hat{\theta}} \mathcal{L}_t(\widehat{\theta_{k'}})$;
  $k' \leftarrow k' + 1$;
**Until** Convergence
**Return** $\hat{\theta}$

---

## 4 PROPOSED METRIC

We introduce a new individual fairness metric based on the following questions: 1) Among the subset of similar individuals, how many pairs have different ground truth outcomes but the same predicted outcomes? And 2) What is the gap of scores resulting from 1) when a GLR is used versus not? Under 1), we aim to quantify the gain in individual fairness through an interventional model such as IFRT and IFDA. Indeed, when a pair of individuals $(i, j)$ are similar, their ground truth labels and predictions can either be the same or different. This results in the following four cases:

a) $y_i = y_j$ and $f(x_i; \hat{\theta}) = f(x_j; \hat{\theta})$
b) $y_i = y_j$ and $f(x_i; \hat{\theta}) \neq f(x_j; \hat{\theta})$
c) $y_i \neq y_j$ and $f(x_i; \hat{\theta}) = f(x_j; \hat{\theta})$
d) $y_i \neq y_j$ and $f(x_i; \hat{\theta}) \neq f(x_j; \hat{\theta})$

While case a) satisfies the individual fairness condition (i.e., similar outcomes for similar individuals) there is no fairness gain (i.e., interventional effect) as individuals in the same ground truth class are assigned in the same predicted class. Cases b) and d) result in unfairness but not gains. Case c) results in a fairness gain as the predicted outcomes of the individuals are the same. Therefore, we define a fairness gain score under c). Such scores can be computed over the entire dataset (global individual fairness score) and for a sub-sample of individuals (local individual fairness score). We consider a general case where the subset of individuals is driven by the similarity matrix $K$ and a user-defined positive threshold $\sigma$ such that the sample size is given by $|K \leq \sigma|$. Thus, we define the unnormalized individual fairness score $I_\alpha(K, \sigma)$ of a model $f$ under a regularization strength $\alpha$ and similarity matrix $K$ as:

$$I_\alpha(K, \sigma) = \frac{|\{i,j \,/\, K_{ij}^U \leq \sigma, y_i \neq y_j \text{ and } f(x_i; \hat{\theta}, \alpha) = f(x_j; \hat{\theta}, \alpha)\}|}{|\{i,j \,/\, K_{ij}^U \leq \sigma, y_i \neq y_j\}|}, \quad (8)$$

As $K$ is symmetric we used the upper triangular matrix $K^U$ in $I_\alpha(K, \sigma)$. A drawback of $I_\alpha(K, \sigma)$ is that the conditions $y_i \neq y_j$ and $f(x_i; \hat{\theta}, \alpha) = f(x_j; \hat{\theta}, \alpha)$ could hold without any regularization leading to false positive or false negative outcomes. Consider a binary classification (e.g., credit allocation), and suppose that the credit allocation for two similar individuals $x_1$ and $x_2$ are different: $y_1 = 1$ and $y_2 = 0$. A true positive outcome on $x_1$ and false positive outcomes on $x_2$ from the model with any regularization ($\alpha = 0$) for instance due to the poor performance of the model will lead to $y_1 \neq y_2$ and $f(x_1; \hat{\theta}, \alpha) = f(x_2; \hat{\theta}, \alpha)$. Inversely, a false negative outcome on $x_1$ and true negative outcomes on $x_2$ will lead to $y_1 \neq y_2$ $and$ $f(x_1; \hat{\theta}, \alpha) = f(x_2; \hat{\theta}, \alpha)$. To account for these cases we define a normalized version of $I_\alpha(K, \sigma)$ as the *Fairness Gain* score *(FG)* as follows:

$$I_\alpha^{norm}(K, \sigma) = I_\alpha(K, \sigma) - I_{\alpha=0}(K, \sigma) \quad (9)$$

From equation (9) we have $I_{\alpha,\beta}^{norm} \in [-1, 1]$ and a model enforces individual fairness if $I_{\alpha,\beta}^{norm}(K, \sigma) > 0$. Given the connection between the pairwise similarity matrix $K$ and the graph Laplacian $L$, $I_\alpha^{norm}(K, \sigma)$ can be extended to $I_\alpha^{norm}(L, \sigma)$.

## 5 EMPRICAL RESULTS

We compared IFRT, IFDA, and Logistic Regression (LR) on the German Credit dataset and used diverse evaluation criteria in addition to the Normalized Fairness Gain. The results are reported in the following.

### 5.1 Datasets and Metrics

We used the German Credit allocation (Credit) dataset, a popular dataset in the algorithmic fairness literature [14] available on the UCI Machine Learning repository obtained from the AI Fairness 360 toolkit [3] to evaluate the models. The data contain $n = 1000$ samples described by $m = 59$ attributes and the task consists of predicting whether a credit should be allocated to an individual ($y = 1$) or not ($y = 0$) based on socio-demographic and credit history data. For data processing details and in-depth descriptions of the data, we refer to [3]. We used the ROC-AUC (AUC), the False Positive Rate (FPR), and the False Negative Rate (FNR) as common metrics in supervised learning and the algorithmic fairness literature [32], and the Prediction Consistency (PC) [36], in addition to the FG and FGN to evaluate the models. We computed PC scores on the Credit dataset by comparing model predictions under original sensitive variables and when these variables were perturbated (through swapping the categories of SEX=Male to SEX=Female (and vice versa), and AGE=Old to AGE=Young (and vice versa)).

### 5.2 Experimental Protocol and Parameter Settings

We ran all algorithms over 200 epochs and used the same initializations (zeros) of $\theta$. Under independent and identical distributions (IID) we used a stratified five-fold cross-validation procedure. We used SEX, AGE, and the two variables (SEX-AGE in Table 1) as sensitive variables under the IID setting. In contrast, under covariate shift, we used AGE=Young from the original variable as the source data, and AGE=Old as the target data and considered only SEX as the sensitive variable. We tested different regularization strength hyperparameter values $\alpha$ and set it to 10 in IFDA and IFRT to balance between the fairness scores and classification scores. For the reweighting vector $W$ we set the values to uniform weights (a vector of ones) under IID and the inverse propensity scores under covariate shift (see [9] for details). We set the similarity threshold values $\sigma_s$ and $\sigma_t$ to 1 (i.e., max $(K)$; which also corresponds to a global IF setting). In all experiments, we used the Euclidean distance as a similarity metric for $K$, $\delta = 5$ as the temperature parameter (i.e., $K_{ij} = e^{-5d(x_i, x_j)}$), the convergence tolerance to $1e^{-7}$ for the training of IFDA and IFRT and $1e^{-10}$ for the inference of IFRT, and learning rates to 0.1 for all methods.

We used a LR model with no penalty as a baseline model. We used the same random seed for LR, IFDA, and IFRT and implemented the LR model with Scikit-Learn. As the method is not a graph Laplacian-based we did not report FG and NFG scores. Throughout the experiments, we noticed that the constrained ($\alpha \neq 0$) and unconstrained ($\alpha = 0$) IFDA and IFRT methods do not have the same number of convergence iterations. Therefore, we computed the FGN at convergence in (9).

### 5.3 Statistical Comparisons

Assuming that the cross-validation scores are normally distributed, we statistically compared the performance of the LR, IFDA, and IFRT models through an analysis of variance (ANOVA)[1]. When the ANOVA null hypothesis (equal mean scores) was rejected, we compared the IFDA and IFRT models using the Tukey honestly significance difference tests (Tukey HSD). The mean and standard deviations of the LR, IFDA, and IFRT models obtained from the five-fold cross-validation under the IID setting and ANOVA (5% significance threshold) are reported in Table 1.

|  | AUC | FNR | FPR | FG | NFG | PC | | |
|---|---|---|---|---|---|---|---|---|
|  |  |  |  |  |  | SEX | AGE | SEX-AGE |
| **LR** | 0.778±0.024 | 0.518±0.04 | 0.13±0.01 | - | - | 0.940±0.01 | 0.920±0.01 | 0.880±0.02 |
| **IFDA** | 0.772±0.02 | 0.763±0.02 | 0.047±0.0 | 0.312±0.01 | 0.027±0.01 | 0.958±0.01 | 0.943±0.01 | 0.915±0.01 |
| **IFRT** | 0.773±0.02 | 0.743±0.03 | 0.053±0.0 | 0.303±0.01 | 0.019±0.01 | 0.963±0.01 | 0.937±0.01 | 0.891±0.01 |
| **ANOVA p-values** | 0.92 | 2.3e-7 | 1.5e-8 | 0.3 | 0.06 | 0.035 | 0.05 | 0.007 |

**Table 1: Mean and standard deviation scores obtained from five-fold cross-validation under the IID setting and ANOVA p-values obtained for LR, IFDA, and IFRT.**

FG and NFG scores in Table 1 can be interpreted as follows: IFDA enforces individual fairness for about 31% of the global sample and IFRT for about 30% (the percentages correspond to FG scores * 100). However, when normalized (NFG scores) these percentages drop to respectively 2.7% and 1.9%.

The ANOVA p-values from Table 1 and post hoc Tukey HSD pairwise comparisons in the Appendix show that the mean NFG scores IFDA, and IFRT are statistically equivalent. For PC scores, when SEX is the sensitive variable, the means of IFRT and LR are statistically different while the means of IFDA and IFRT, and the means of IFDA and LR are statistically the same. For the sensitive variables AGE and SEX-AGE, the means of IFDA and LR are statistically different while IFDA and IFRT, and IFRT and LR have the same statistical means.

Overall, the experimental results show that IFDA and IFRT achieved similar performances on all evaluation metrics. For some sensitive variables, IFDA, IFRT, and LR achieved similar PC performances indicating that an AI model without any fairness regularization such as the LR model can perform similarly to fairness-enhanced models such as IFDA and IFRT. Furthermore, these results show that PC scores can be misleading in measuring individual fairness as AI models might seem to optimize this metric while in reality, the fairness gain is small as indicated by FGN scores in Table 1.

---
[1] As we obtained similar cross-validation standard deviations for all models (see Table 2) the homoscedasticity criterion was met.

**5.4 Results under the covariate shift setting**

Figure 1 describes the results of experiments obtained under the covariate shift setting. The x-axes refer to the number of iterations/epochs and the y-axes to the scores. The figure shows a positive relation between the number of iterations and FNR, FG, and NFG for IFRT and FPR for IDFA. As observed under the IID setting, there is a contrast between PC scores and NFG scores. We did not statistically compare the performance of IFDA and IFRT under the covariate shift setting but expect different statistical performances.

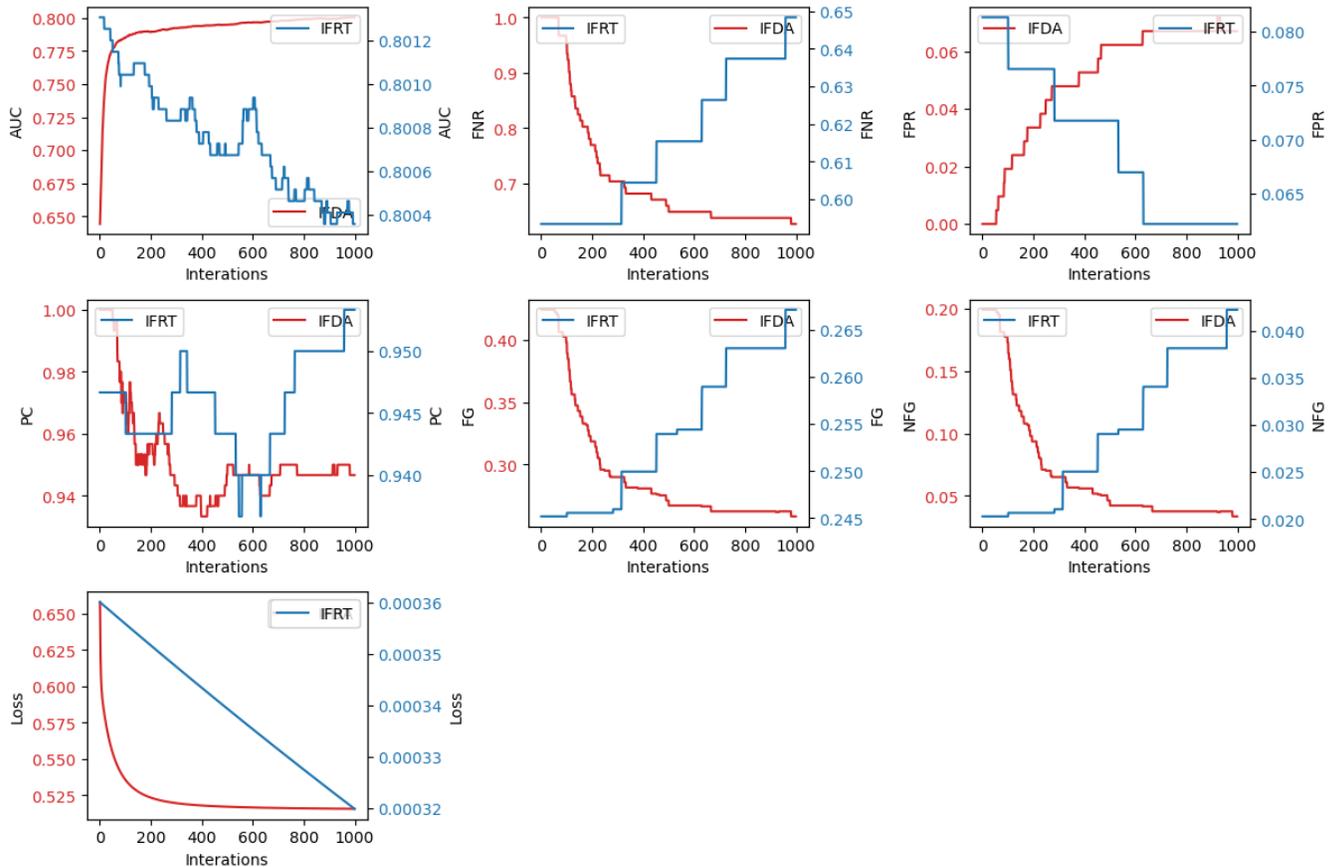

Figure 1: Target scores obtained under the covariate shift setting

**6 LIMITATIONS**

While our study provides insights into in-processing individual fairness methods and metrics effectiveness, it has three main drawbacks. First, we performed the experiments using only one small dataset. The interpretations of our results should then be restricted to this data. We tried to include more datasets in the analysis, however, the pairwise similarity matrix is computationally highly demanding. For instance, we could not run on a 16 GB RAM i7-1355U 1.70 GHz Dell Laptop the pairwise similarity matrix computation of the Bank Marketing dataset from the UCI Machine Learning repository [3]. The data has 30488 samples and 58 variables. Even other small datasets such as the Adult and COMPAS [3] result in a memory error or a while to run. This limits the practicality of graph Laplacian-based methods for individual fairness enforcement. Nevertheless, the results on the Credit dataset could inform the pitfalls of PC scores and the effectiveness of graph Laplacian-based fairness constraints. Second, the individual fairness framework studied in this work relies heavily on the similarity between individuals. Traditional metrics such as the Euclidean may be unfair under certain applications which constitute another barrier. However, there is an increasing interest in developing or learning fair similarity measures from the data, mostly for natural language applications [23].

## 7 CONCLUSION

We introduced a new graph Laplacian-based individual fairness technique and metric in this paper. The new model extends IFDA and TFCS by overcoming their limitation respectively by requiring access to the target data during the training phase and access to sensitive variables to enforce fairness. The proposed model promotes individual fairness in two steps: first on the source data, and then on the target data by tuning the model. Moreover, the new model can be applied under IID and covariate shift settings through instance reweighting. We showed through the experiments that some fairness and classification metrics should be picked and interpreted with particular attention as a model can attain good performances on some metrics (e.g., PC) while providing a little fairness gain. We noted that the current work is limited by the diversity of data and optimal parameter setting. In follow-up work, we will investigate effective ways to compute the graph Laplacian matrix.

## LIST OF ABBREVIATIONS

AI: Artificial Intelligence  
GLR: Graph Laplacian Regularizer  
PC: Prediction Consistency  
IFDA: Individual Fairness through Domain Adaptation  
IFRT: Individual Fairness through Reweighting and Tuning  
FG: Fairness Gain  
FGN: Normalized Fairness Gain  
AUC: Area under the ROC curve  
FNR: False Negative Rate  
FPR: False Positive Rate  
IID: Independent and Identically Distributed  
ANOVA: Analysis of Variance

**APPENDIX**

This appendix presents the pairwise statistical analysis results obtained from the Tukey HSD tests. In all tables, IFDA, IFRT, and LR models are respectively denoted by 0, 1, and 2.

```
Tukey's HSD Pairwise Group Comparisons of FNR Scores (95.0% Confidence Interval
)
Comparison  Statistic  p-value  Lower CI  Upper CI
 (0 - 1)       0.020    0.668    -0.041     0.081
 (0 - 2)       0.245    0.000     0.184     0.307
 (1 - 0)      -0.020    0.668    -0.081     0.041
 (1 - 2)       0.225    0.000     0.164     0.287
 (2 - 0)      -0.245    0.000    -0.307    -0.184
 (2 - 1)      -0.225    0.000    -0.287    -0.164

Tukey's HSD Pairwise Group Comparisons of FPR Scores (95.0% Confidence Interval
)
Comparison  Statistic  p-value  Lower CI  Upper CI
 (0 - 1)      -0.006    0.630    -0.022     0.011
 (0 - 2)      -0.083    0.000    -0.099    -0.067
 (1 - 0)       0.006    0.630    -0.011     0.022
 (1 - 2)      -0.077    0.000    -0.093    -0.061
 (2 - 0)       0.083    0.000     0.067     0.099
 (2 - 1)       0.077    0.000     0.061     0.093

Tukey's HSD Pairwise Group Comparisons of PC Scores with SEX as the sensitive
variable(95.0% Confidence Interval)
Comparison  Statistic  p-value  Lower CI  Upper CI
 (0 - 1)      -0.005    0.813    -0.027     0.017
 (0 - 2)       0.018    0.107    -0.004     0.040
 (1 - 0)       0.005    0.813    -0.017     0.027
 (1 - 2)       0.023    0.037     0.001     0.045
 (2 - 0)      -0.018    0.107    -0.040     0.004
 (2 - 1)      -0.023    0.037    -0.045    -0.001

Tukey's HSD Pairwise Group Comparisons of PC Scores with AGE as the sensitive
variable(95.0% Confidence Interval)
Comparison  Statistic  p-value  Lower CI  Upper CI
 (0 - 1)       0.006    0.769    -0.017     0.029
 (0 - 2)       0.023    0.049     0.000     0.046
 (1 - 0)      -0.006    0.769    -0.029     0.017
 (1 - 2)       0.017    0.159    -0.006     0.040
 (2 - 0)      -0.023    0.049    -0.046    -0.000
 (2 - 1)      -0.017    0.159    -0.040     0.006

Tukey's HSD Pairwise Group Comparisons of PC Scores with SEX-AGE as the sensiti
ve variable(95.0% Confidence Interval)
Comparison  Statistic  p-value  Lower CI  Upper CI
 (0 - 1)       0.024    0.055    -0.000     0.048
 (0 - 2)       0.035    0.006     0.011     0.059
 (1 - 0)      -0.024    0.055    -0.048     0.000
 (1 - 2)       0.011    0.475    -0.013     0.035
 (2 - 0)      -0.035    0.006    -0.059    -0.011
 (2 - 1)      -0.011    0.475    -0.035     0.013
```